\documentclass[10pt]{IEEEtran}

\usepackage{amsmath,amssymb}
\usepackage[ruled]{algorithm2e}
\usepackage{graphicx,epsfig,url}
\usepackage[caption=false,font=footnotesize]{subfig}
\usepackage[usenames, dvipsnames]{color}  
\usepackage{multirow}
\usepackage{cite}
\usepackage{color}

\def\A{\mathbf{A}}
\def\B{\mathbf{B}}

\def\K{\mathbf{K}}

\def\u{\boldsymbol{u}}
\def\v{\boldsymbol{v}}
\def\w{\boldsymbol{w}}
\def\x{\boldsymbol{x}}
\def\z{\boldsymbol{z}}

\def\y{\boldsymbol{y}}

\def\p{\boldsymbol{p}}
\def\f{\boldsymbol{f}}
\def\g{\boldsymbol{g}}
\def\h{\boldsymbol{h}}
\def\r{\boldsymbol{r}}
\def\amu{\boldsymbol{\mu}} 

\ifCLASSINFOpdf
\else
\fi

\newtheorem{theorem}{Theorem}
\newtheorem{proposition}[theorem]{Proposition}

\begin{document}

\markboth{Submitted to IEEE Signal Processing Letters}{}

\title{Fast High-Dimensional Kernel Filtering}

\author{Pravin Nair,~\IEEEmembership{Student~Member,~IEEE} and Kunal N. Chaudhury,~\IEEEmembership{Senior~Member,~IEEE} 
 }

\maketitle

\begin{abstract}
The bilateral and nonlocal means filters are instances of kernel-based filters that are popularly used in image processing. It was recently shown that fast and accurate bilateral filtering of grayscale images can be performed using a low-rank approximation of the kernel matrix. More specifically, based on the eigendecomposition of the kernel matrix, the overall filtering was approximated using spatial convolutions, for which efficient algorithms are available. Unfortunately, this technique cannot be scaled to high-dimensional data such as color and hyperspectral images. This is simply because one needs to compute/store a large matrix and perform its eigendecomposition in this case. We show how this problem can be solved using the Nystr$\ddot{\text{o}}$m method, which is generally used for approximating the eigendecomposition of large matrices. The resulting algorithm can also be used for nonlocal means filtering. We demonstrate the effectiveness of our proposal for bilateral and nonlocal means filtering of color and hyperspectral images. In particular, our method is shown to be competitive with state-of-the-art fast algorithms, and moreover it comes with a theoretical guarantee on the approximation error. 
 \end{abstract}
 
\begin{IEEEkeywords}
Kernel Filter, Nystr$\ddot{\text{o}}$m Method, Approximation, Fast Algorithm, Error Bound.
\end{IEEEkeywords}

\section{Introduction}

The bilateral and nonlocal means filters \cite{tomasi1998bilateral,buades2005non} are widely used for edge-preserving smoothing and denoising of images \cite{paris2009bilateral,milanfar2013tour}. These are instances of kernel filters, where the similarity (affinity) between pixels is measured using a symmetric kernel. We refer the reader to \cite{milanfar2013tour} for an excellent review of kernel filters. While they have proven to be useful in practice, a flip side of kernel filtering, including bilateral filtering (BLF) and nonlocal means (NLM), is their computational complexity \cite{paris2009bilateral}. Nevertheless, several fast algorithms have been proposed, e.g. \cite{durand2002fast,paris2006fast,chen2007real,Porikli2008,Yang2009,Chaudhury2011,Kamata2015,Yang2015,Chaudhury2016,Ghosh2016,Sugimoto2016,ipol2017184,papari2017fast,adams2009gaussian,adams2010fast,gastal2012adaptive,mozerov2015global,pravin2017filtering,mahmoudi2005,wang2006,darbon2008,ghosh2016nlm}, which can speed up BLF and NLM, without compromising their filtering quality. See \cite{mozerov2015global,Kamata2015,ghosh2016nlm} for a survey of these algorithms.
Unfortunately, most algorithms only work with grayscale images, and cannot be extended to color, multispectral, and hyperspectral images. 

Algorithms for fast BLF of color images have been proposed in \cite{Yang2015,mozerov2015global,ghosh2016fast,tu2016constant,sugimoto2016fast}. However, to the best of our knowledge, these methods have not been extended for multispectral and hyperspectral images. 
Fast algorithms for generic high-dimensional BLF and NLM have been proposed in \cite{adams2009gaussian,adams2010fast,gastal2012adaptive,karam2018monte}. A common feature of these algorithms is that they use data clustering or tessellation in high-dimensions. The state-of-the-art fast algorithms for color BLF are \cite{adams2010fast,mozerov2015global}, and for color NLM is \cite{gastal2012adaptive}.

More recently, it was shown in \cite{Sugimoto2016,papari2017fast} that fast BLF of grayscale images can be performed using 
the partial eigendecomposition of the kernel matrix. In fact, the interpretation of BLF (and NLM) as kernel filters goes back to  \cite{talebi2014global,talebi2014nonlocal,talebi2016asymptotic}. 
While the Nystr$\ddot{\text{o}}$m method has widely been used in machine learning \cite{williams01usingthe,fowlkes2004spectral,zhang2008improved}, it appears that \cite{talebi2014global} is the first to apply this for image filtering. 
Note that, unlike \cite{Sugimoto2016,papari2017fast}, the spatial and range kernel are treated as a single kernel in \cite{talebi2014global,talebi2014nonlocal,talebi2016asymptotic}.

The differences between our and related approaches are:

\indent $\bullet$ As explained in detail in \S \ref{Background}, it is difficult to scale \cite{Sugimoto2016,papari2017fast} for filtering high-dimensional (even color) images, since one needs to populate a huge kernel matrix and compute its eigendecomposition. 
We propose to use the Nystr$\ddot{\text{o}}$m method to solve this problem. As a result, we are able to perform BLF and NLM of color and hyperspectral images. 

\indent $\bullet$ The first difference with \cite{talebi2014global,talebi2014nonlocal,talebi2016asymptotic} is that we use clustering instead of uniform sampling for the Nystr$\ddot{\text{o}}$m approximation. A significant improvement in filtering accuracy is achieved as a result. The other difference is that if a spatial kernel has to incorporated in  \cite{talebi2014global,talebi2014nonlocal,talebi2016asymptotic}, then the Nystr$\ddot{\text{o}}$m approximation needs to be performed  in the spatio-range space. 
However, we handle the spatial and range components differently---fast convolutions are used for the spatial component and Nystr$\ddot{\text{o}}$m approximation is used for the range component.
As a result, we require lesser samples for the Nystr$\ddot{\text{o}}$m approximation.

\indent $\bullet$ In \cite{tu2016constant,sugimoto2016fast},  clustering is used to compute ``intermediate'' images, which are interpolated to get the final output. On the other hand, clustering is used in our method just to obtain the ``landmark points'' for the Nystr$\ddot{\text{o}}$m approximation. 

\indent $\bullet$ Compared to \cite{adams2009gaussian,adams2010fast,gastal2012adaptive,mozerov2015global}, our algorithm is conceptually simple and easy to implement. Moreover, we are able to derive a bound on the filtering error incurred by the approximation. Such a guarantee is not offered by  \cite{adams2009gaussian,adams2010fast,gastal2012adaptive,mozerov2015global}.

The rest of the paper is organized as follows. In \S \ref{Background}, we introduce the notion of kernel filtering, and explain the core problem in relation to the spectral approximations in \cite{Sugimoto2016,papari2017fast}. We use the Nystr$\ddot{\text{o}}$m method in \S \ref{Proposed} to overcome  this problem. Numerical results are reported in \S \ref{Numerical} and we conclude in \S \ref{Conc}.

\section{Background}
\label{Background}

We begin by formulating BLF and NLM as kernel filters \cite{milanfar2013tour}.
Suppose the input image is $\f: \Omega  \to [0,R]^n$, where $\Omega \subset \mathbb{Z}^{d}$ is the spatial domain, $[0,R]^n$ is the  range space, and $d$ (resp. $n$) is the dimension of the domain (resp. range). Let $\p : \Omega  \to [0,R]^{\rho}$ be the \textit{guide} image, which is used to control the filtering. For standard BLF, $\f$ and $\p$ are identical, and  $n=\rho=1$ and $3$ for grayscale and color images. However, $\f$ and $\p$ (also $n$ and $\rho$) can be different for joint BLF \cite{paris2009bilateral}. For NLM, $\rho$ is generally larger than $n$, where $\rho$ is the number of pixels in a \textit{patch} \cite{buades2005non}. Let $\kappa : \mathbb{R}^{\rho} \times \mathbb{R}^{\rho} \to \mathbb{R}$ be the \textit{range} kernel. The filtered output $\g: \Omega  \to [0,R]^n$ is given by
\begin{equation}
\label{num}
\g(\x)=\frac{\sum_{\y \in W_{\x}} \omega(\x-\y)  \kappa\big(\p(\x),\p(\y)\big)  \f(\y)}{\sum_{\y \in W_{\x}} \omega(\x-\y)  \kappa\big(\p(\x),\p(\y)\big)},
\end{equation}
where $W_{\x}$ is a square window around $\x \in \Omega$ consisting of $(2S+1)^d$ pixels, with $S$ being the window radius. 
The \textit{spatial} kernel $\omega: \mathbb{Z}^d \to \mathbb{R}$ controls the weighting of the neighboring pixels involved in the averaging. At this point, we just assume that $\kappa$ is symmetric, i.e., $\kappa(\boldsymbol{t},\boldsymbol{s})=\kappa(\boldsymbol{s},\boldsymbol{t})$ for $\boldsymbol{t},\boldsymbol{s} \in \mathbb{R}^{\rho}$. For example, $\kappa(\boldsymbol{t},\boldsymbol{s}) = \exp(- \theta \lVert \boldsymbol{s}- \boldsymbol{t}\rVert^2), \theta > 0$, for standard BLF and NLM, where $\lVert \cdot \rVert$ is the Euclidean norm.

It was shown in \cite{Sugimoto2016,papari2017fast} that the non-linear operations in \eqref{num} can be computed using convolutions by approximating $\kappa$. For convenience, we will describe this using our notations. Let the actual range of $\p$ be 
\begin{equation}
\label{range}
\mathfrak{R}  = \big\{\p(\x) : \x \in \Omega \big\}.
\end{equation}
We emphasize that $\mathfrak{R}$ is a list and not a set, i.e., we allow repetition of elements in $\mathfrak{R}$. In particular, let $\mathfrak{R} =\{\r_1,\r_2,....,\r_m\}$ be some ordering of the elements in $\mathfrak{R}$, where $m$ is the number of elements. This means that, given $\ell \in [1,m]$, $\r_{\ell}=\p(\x)$ for some $\x \in \Omega$. 
We track this correspondence using the \textit{index map} $\iota: \Omega  \to [1,m]$, where 
\begin{equation}
\label{indexdef}
\iota(\x)=\ell \qquad \text{if} \ r_{\ell}=p(\x).  
\end{equation}
We next define the kernel matrix $\K \in \mathbb{R}^{m \times m}$ given by
\begin{equation}
\label{kernel}
\K(i,j) = \kappa(\r_i, \r_j).
\end{equation}
In terms of \eqref{kernel}, we can write \eqref{num} as
\begin{equation}
 \label{num1}
 \g(\x)=\frac{\sum_{\y \in W_{\x}} \omega(\x-\y)  \K\big(\iota(\x),\iota(\y)\big)  \f(\y)}{\sum_{\y \in W_{\x}} \omega(\x-\y)  \K\big(\iota(\x),\iota(\y)\big)}
\end{equation}
It is clear from \eqref{kernel} that $\K$ is symmetric. In particular, let the eigendecomposition of $\K$ be
\begin{equation}
\label{eigdecom}
\K = \sum_{k=1}^{m} \lambda_k \u_k \u_k^{\top},
\end{equation}
where $\lambda_1,\ldots, \lambda_m \in \mathbb{R}$ are its eigenvalues, and $\u_1,\ldots,\u_m \in \mathbb{R}^m$ are the corresponding eigenvectors. Substituting \eqref{eigdecom} in \eqref{num1}, we can write its numerator as
\begin{equation*}
\sum_{\y \in W_{\x}} \omega(\x-\y)   \left\{  \sum_{k=1}^{m} \lambda_k \u_k \big(\iota(\x) \big) \u_k \big(\iota(\y) \big)\right\} \f(\y).
\end{equation*}
On switching the sums, this becomes
\begin{equation}
\label{switch}
 \sum_{k=1}^{m} \lambda_k \u_k \big(\iota(\x) \big) (\omega \ast \boldsymbol{h}_k)(\x),
\end{equation}
where $\omega \ast \boldsymbol{h}_k$ denotes the convolution of the image $\boldsymbol{h}_k(\x) = \u_k \big(\iota(\x) \big) \f(\x)$ with $\omega$. An identical argument applies for the denominator. In summary, we can compute \eqref{num1} using convolutions, for which several efficient algorithms are available \cite{young1995recursive,sugimoto2013fast}. Moreover, by considering just the largest eigenvalues, fast and accurate approximations can be obtained \cite{Sugimoto2016,papari2017fast}. 

Unfortunately, computing the full kernel and its eigendecomposition becomes prohibitively expensive when $\rho$ is large. Just as an example, consider an $8$-bit color image for which $R=255$ and $\rho=3$. Even if we assume that $m$ is just $10\%$ of the maximum range cardinality ($=256^3$), we will still need to populate a $3 \text{ million} \times 3 \text{ million}$ matrix, and compute its eigenvalues. The situation is worse for hyperspectral images, where $\rho$ is of the order of tens, or even hundreds.

\section{Proposed Method}
\label{Proposed}

Originally, the Nystr$\ddot{\text{o}}$m method was used for approximating the solution of functional eigenvalue problems \cite{nystrom1930praktische,baker1977numerical}. The method has found useful applications in machine learning and computer vision for approximating the eigendecomposition of large matrices \cite{williams01usingthe,fowlkes2004spectral,talebi2014global}. In the present context, the goal is to approximate 
\eqref{eigdecom} using a decomposition of the form
\begin{equation}
\label{hatK}
\widehat{\K}=\sum_{k=1}^{m_0} \alpha_k \v_k \v_k^\top,
\end{equation}
where $\alpha_k \in \mathbb{R}$ and $\v_k \in \mathbb{R}^{m}$. Clearly, the rank of $\widehat{\K}$ is at most $m_0$. Thus, for small $m_0$, $\widehat{\K}$ is a low-rank approximation of $\K$. A large $m_0$ results in a better approximation, but at higher computational cost. In practice, a good tradeoff is required.

The original kernel $\K$ is defined on $\mathfrak{R}$. In the Nystr$\ddot{\text{o}}$m method \cite{nystrom1930praktische,baker1977numerical}, we first construct a smaller kernel $\A$, compute its eigendecomposition, and then ``extrapolate'' the eigenvectors of $\A$ to approximate those of $\K$. More precisely, we pick few \textit{landmarks} points from $\mathfrak{R}$, say, $\mathfrak{R}_0 = \{\amu_1,\ldots,\amu_{m_0}\}$, and define a kernel $\A \in \mathbb{R}^{m_0 \times m_0}$ on $\mathfrak{R}_0$:
\begin{equation}
\label{defA}
\A(i,j) = \kappa(\amu_i, \amu_j) \qquad \big(i,j \in [1,m_0] \big).
\end{equation}
Clearly, $\A$ is symmetric, and its size is much smaller than $\K$. Thus, we can efficiently compute its eigendecomposition:
\begin{equation}
\label{eigA}
\A=\sum_{k=1}^{m_0} \alpha_k \w_k \w_k^\top,
\end{equation}
where $\alpha_k \in \mathbb{R}$ and $\w_k \in \mathbb{R}^{m_0}$. We next construct $\B \in \mathbb{R}^{m_0 \times m}$ on $\mathfrak{R}_0 \times \mathfrak{R}$ given by
\begin{equation}
\label{defB}
\B(i,j) = \kappa(\amu_i, \r_j),
\end{equation}
where $i \in [1,m_0]$ and $j \in [1,m]$. This captures the kernel values between the points in $\mathfrak{R}$ and the landmark points. This matrix is used to extrapolate $\w_k$ as follows:  
\begin{equation}
\label{vk}
\v_k = \frac{1}{\alpha_k} \B^\top \!\w_k \qquad \big(k \in [1,m_0] \big).
\end{equation}
This completes the specification of $\alpha_k$ and $\v_k$ in \eqref{hatK}. We refer the reader to \cite{fowlkes2004spectral} for the intuition behind the approximation. 
The effective speedup of replacing \eqref{eigdecom} by \eqref{hatK} is $\mathcal{O}(m/m_0)^3$. This is because the complexity of eigendecomposition of a $k\times k$ matrix is $\mathcal{O}(k^3)$ \cite{pan1999complexity}. In particular, the speedup is significant since $m_0\ll m$. 
As will be evident shortly, we just need to compute $(\alpha_k)$
and $(\v_k)$; we will not  use $\widehat{\K}$ explicitly.

Following \cite{zhang2008improved}, we select the landmark points by clustering $\mathfrak{R}$. More specifically, we partition $\mathfrak{R}$ into $m_0$ disjoint sets using $k$-means clustering, and take the centroids to be the landmarks. Note that, though $\mathfrak{R}_0$ is not guaranteed to be a subset of $\mathfrak{R}$, we can still apply the above approximation. 

It was shown in \cite{zhang2008improved} that the \textit{kernel error} can be bounded by the \textit{quantization error}. More specifically, let  $\Vert \K - \widehat\K \rVert_{\text{F}}$ be the kernel error ($\lVert \cdot \rVert_{\text{F}}$ is the Frobenius norm), 
and let 
\begin{equation*}
e =  \sum_{i=1}^m \lVert \boldsymbol{r}_i - \amu_{c(i)} \rVert^2
\end{equation*}
be the quantization error, where $c(i)$ is the minimizer of $\lVert \boldsymbol{r}_i - \amu_j \rVert$ over $j \in [1,m_0]$. Then the following bound holds  \cite{zhang2008improved}.
\begin{proposition}
\label{proposition1} \textit{Suppose there exists some $L>0$  such that, for $\w,\x,\y,\z \in \mathfrak{R}$, 
\begin{equation*}
{\big(\kappa(\x,\y) - \kappa(\w,\z)\big)}^2 \leq L \big({\lVert(\x-\w)\rVert}^2 + {\lVert(\y-\z)\rVert}^2\big).
\end{equation*}
Then the approximation error can be bounded as  
\begin{equation}
\label{Nystrombound}
\Vert \K - \widehat\K \rVert_{\mathrm{F}} \leq c_1 \sqrt{e} + c_2 e,  
\end{equation}
where the positive constants $c_1$ and $c_2$ do not depend on $e$. In particular, \eqref{Nystrombound} holds when $\kappa$ is a Gaussian.}
\end{proposition}

Proposition \ref{proposition1} suggests that we can reduce the kernel error by making $e$ small. 
However, $e$ measures how well $\Theta$ is represented by the landmark points.
Following this observation,  $k$-means clustering was used in \cite{zhang2008improved} for determining the landmarks. It was empirically shown in \cite{zhang2008improved}  that clustering indeed results in smaller error over uniform sampling \cite{fowlkes2004spectral,talebi2014global}. 
We will see that this is also true for our algorithm.

We arrive at a fast algorithm by replacing $\K$  by $\widehat{\K}$. 
It is clear from \eqref{switch} that the resulting approximation is given by
\begin{equation}
\label{finalnum}
\hat\g(\x)  = \frac{1}{\hat\eta(\x) }\sum_{k=1}^{m_0} \alpha_k \v_k \big(\iota(\x) \big) (\omega \ast \boldsymbol{h}_k)(\x),
\end{equation}
\begin{equation}
\label{finalden}
\hat\eta(\x)  = \sum_{k=1}^{m_0} \alpha_k \v_k \big(\iota(\x) \big) (\omega \ast d_k)(\x),
\end{equation}
where  $d_k: \Omega \to \mathbb{R}$ and $\boldsymbol{h}_k: \Omega \to \mathbb{R}^n$ are defined as $d_k(\x) = \v_k (\iota(\x) )$ and $\boldsymbol{h}_k(\x) = d_k(\x) \f(\x)$.

The computation of \eqref{finalnum} and \eqref{finalden} involves $(n+1)m_0$ convolutions, since for each $k \in [1,m_0]$, there are $n$ convolutions in \eqref{finalnum} and one in \eqref{finalden}. The main point is that we have been able to express the non-linear kernel filter using convolutions, for which efficient algorithms are available. In particular, \eqref{finalnum} and \eqref{finalden} can be performed using $\mathcal{O}(1)$ operations (w.r.t. the size of the spatial kernel), when $\omega$ is a box or Gaussian \cite{deriche1993recursively,young1995recursive,sugimoto2013fast}. 
The overall algorithm is described in Algorithm \ref{fastalgo} (source code in \cite{Sourcecode}), where the symbols $\oplus,\otimes$ and $\oslash$ are used to denote pixelwise addition, multiplication, and division. 
The complexity of $k$-means clustering and the eigendecomposition of $\A$ are $\mathcal{O}(|\Omega| m_0 \rho)$ \cite{Tan2005}  and $\mathcal{O}({m_0}^3)$ \cite{pan1999complexity}.
On the other hand, the complexity of the convolutions  in \eqref{finalnum} and \eqref{finalden} is $\mathcal{O}(|\Omega| m_0(n+\rho))$, where $|\Omega|$ is the number of pixels. 
Since the complexity of the brute-force implementation is $\mathcal{O}\big(|\Omega| (2S+1)^d(n+\rho)\big)$ \cite{paris2009bilateral}, and convolutions are the dominant operations in our algorithm, we obtain an effective speedup of $(2S+1)^d/m_0$. 
This is significant as $S$ is typically large \cite{paris2009bilateral}. 

\begin{algorithm}
 \textbf{Input}: $\f : \Omega \to \mathbb{R}^n$ and $\p : \Omega \to \mathbb{R}^\rho$, kernels $\omega$ and $\kappa$\;
\textbf{Parameter}: Number of landmarks $m_0$\;
\textbf{Output}: Approximation in \eqref{finalnum}\;
Form $\mathfrak{R}$ in \eqref{range} and index map $\iota$ in \eqref{indexdef}\;
$\{\amu_k\} \leftarrow$ partition $\mathfrak{R}$ into $m_0$ clusters using $k$-means\; \label{cluster}
Construct ${\A}$ and $\B$ in \eqref{defA} and \eqref{defB} using $\kappa$ and $\p$\; 
Compute the eigendecomposition of $\A$ in \eqref{eigA}\;
Initialize $ \boldsymbol{\zeta}: \Omega \to \mathbb{R}^n$ and $\eta: \Omega \to \mathbb{R}$ with zeros\;
\For{$k=1,\ldots,m_0$}{
$\v_k = (1/\alpha_k) \B^{\top}\! \w_k$\; 
\For{$\x \in \Omega$}{
$d_k(\x) = \v_k(\iota(\x))$\; 
$\h_k(\x) =d_k(\x)  \f(\x)$\; 
}
$ \boldsymbol{\zeta} \leftarrow \boldsymbol{\zeta} \oplus \big(\alpha_k \cdot d_k \otimes(\omega \ast \h_k)\big)$\; \label{conv1}
$\eta \leftarrow \eta \oplus \big(\alpha_k \cdot d_k \otimes (\omega \ast d_k)\big)$\; \label{conv2}
}
${\hat\g} \leftarrow \boldsymbol{\zeta} \oslash \eta$.
\caption{Fast Kernel Filtering.}
\label{fastalgo}  
\end{algorithm}

We now comment on the filtering accuracy, namely, how well  is \eqref{num} approximated by \eqref{finalnum}. Intuitively, we expect  the approximation to be accurate if $\widehat{\K} \approx \K$. In fact, since the difference $\Vert \K - \widehat\K \rVert_{\mathrm{F}}$ is controlled by the quantization error (Proposition \ref{proposition1}), we have the following result.
\begin{theorem}
\label{theorem1} \textit{Suppose $\omega$ and $\kappa$ are positive, and $\kappa$ satisfies the property in Proposition \ref{proposition1}. Then
\begin{equation}
 \label{bound}
\|\hat\g-\g\|_\infty = \max_{\x \in \Omega} \  \lVert \hat\g(\x) - \g(\x) \rVert  \leq C_1 \sqrt{e} + C_2 e,
\end{equation}
where $C_1,C_2 >0$ do not depend on $e$.}
\end{theorem}
The main steps of the derivation are given in the supplement. Theorem \ref{theorem1} is true for BLF and NLM, where $\kappa$ is a Gaussian. A practical implication of this result is that the filtering accuracy is guaranteed to increase with $m_0$ (Figure $4$ in the supplement). 
Deriving a similar bound is difficult for \cite{adams2009gaussian,adams2010fast,gastal2012adaptive,mozerov2015global}.

\section{Results}
\label{Numerical}
       We demonstrate the effectiveness of our algorithm for BLF and NLM of high-dimensional images 
by comparing it with state-of-the-art algorithms.          
Instead of standard NLM \cite{buades2005non}, we have used PCA-NLM \cite{tasdizen2009principal}, where the denoising performance of the former is improved by applying PCA on the collection of patches. As for the dataset, we have used the color images from \cite{ImageSource2} and the hyperspectral images from \cite{ImageSource4}.        
Experiments were performed using Matlab on a $3.4$ GHz quad-core machine with $32$ GB memory. 
The spatial kernel $\omega$ for BLF  is a Gaussian (covariance $\sigma^2 \mathbf{I}$ and $S=3\sigma$), while it is a box in PCA-NLM. The range kernel $\kappa$ is Gaussian (covariance $\theta^2 \mathbf{I}$) for both BLF and PCA-NLM.
We have used the fast $\mathcal{O}(1)$ algorithm in \cite{young1995recursive} when $\omega$ is a Gaussian, and the Matlab routine ``imfilter'' when $\omega$ is a box. Note that we can also use other fast Gaussian filters \cite{deriche1993recursively,sugimoto2013fast} if higher accuracy is desired. 

\begin{figure*}
\centering
\subfloat[Clean/Noisy ($20$ dB).]{\includegraphics[width=0.19\linewidth,height=0.18\linewidth]{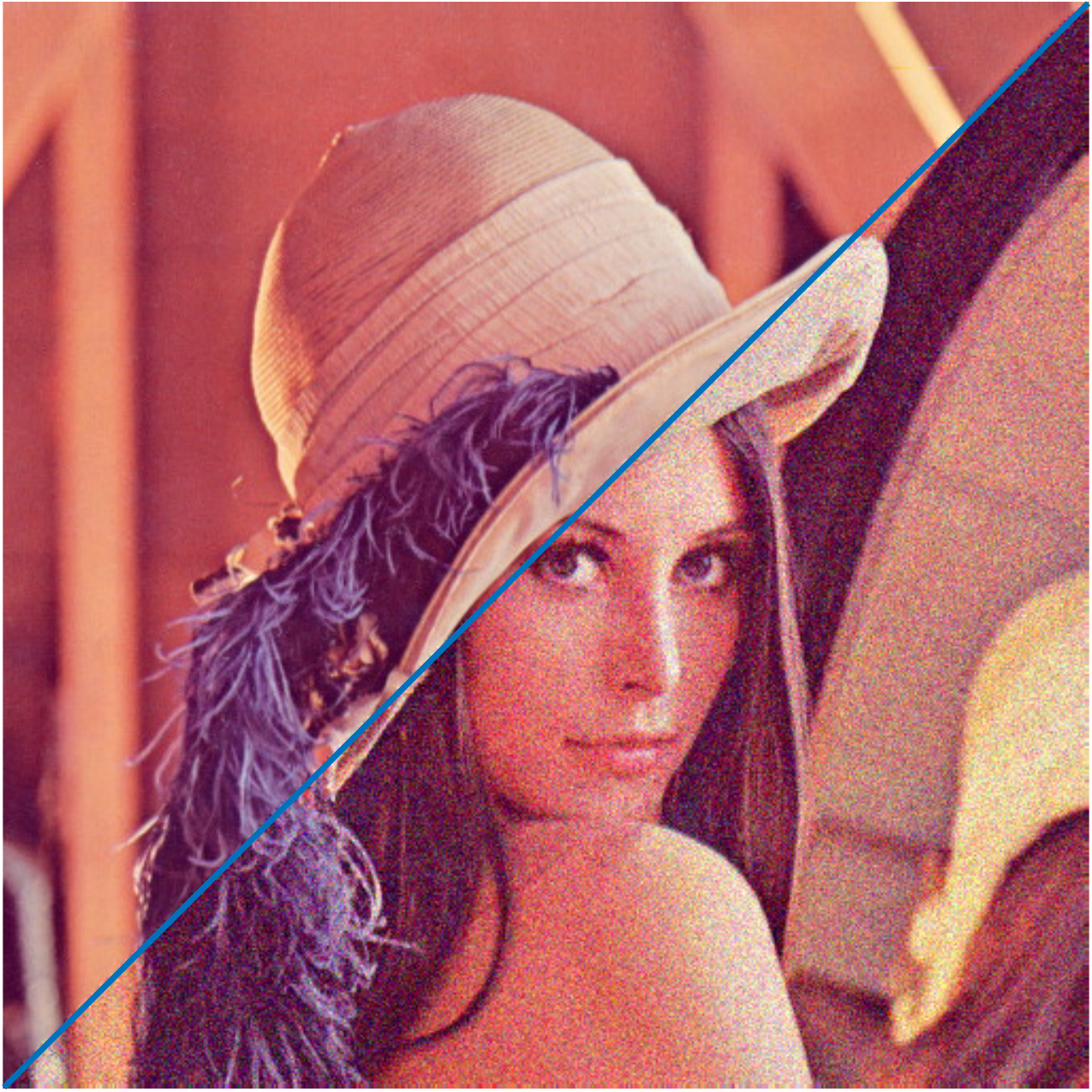}} \hspace{0.5mm}
\subfloat[ ($420$, $30.2$, $0.88$).]{\includegraphics[width=0.19\linewidth,height=0.18\linewidth]{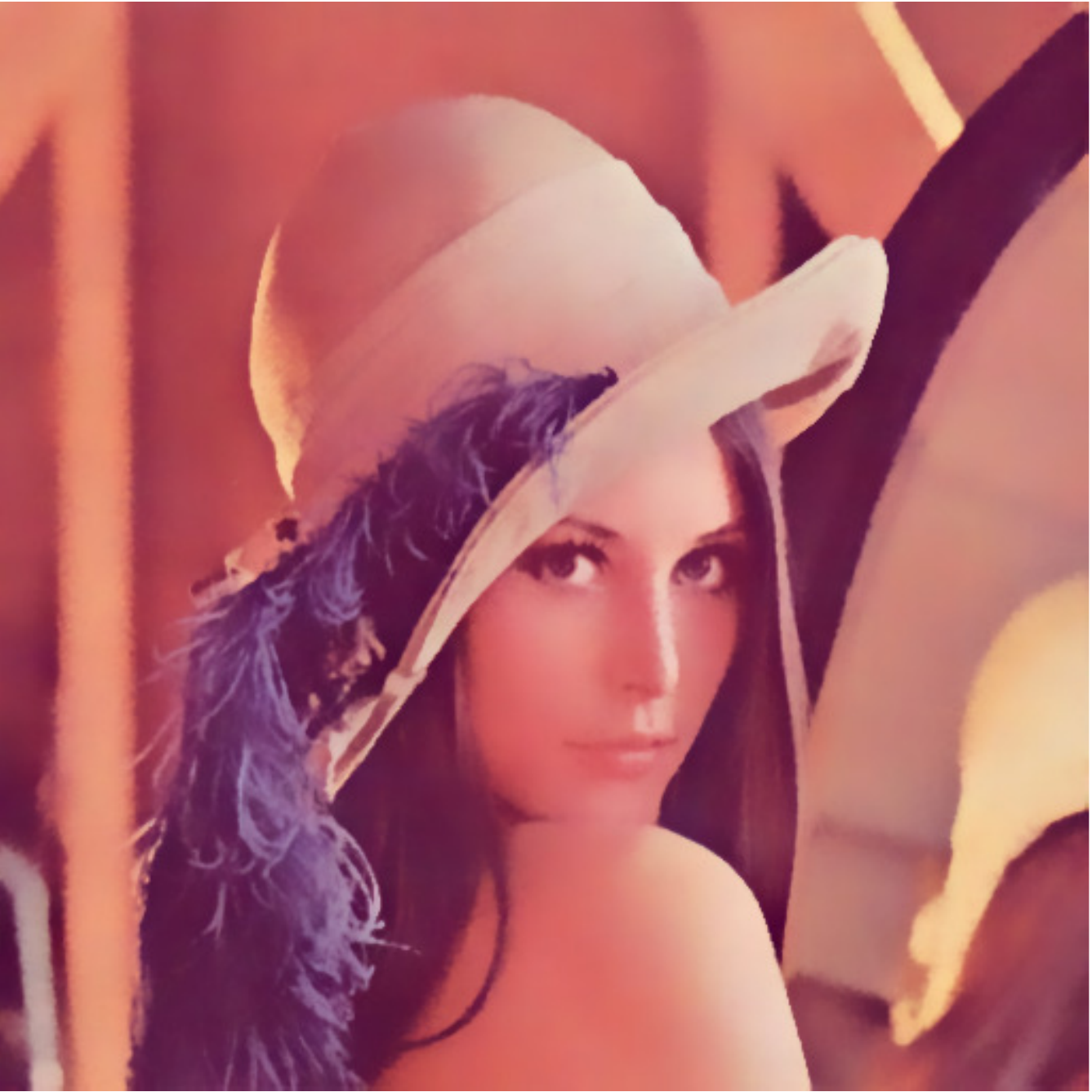}} \hspace{0.5mm}
\subfloat[\textbf{Ours} ($2.6$, $30.1$, $0.88$).]{\includegraphics[width=0.19\linewidth,height=0.18\linewidth]{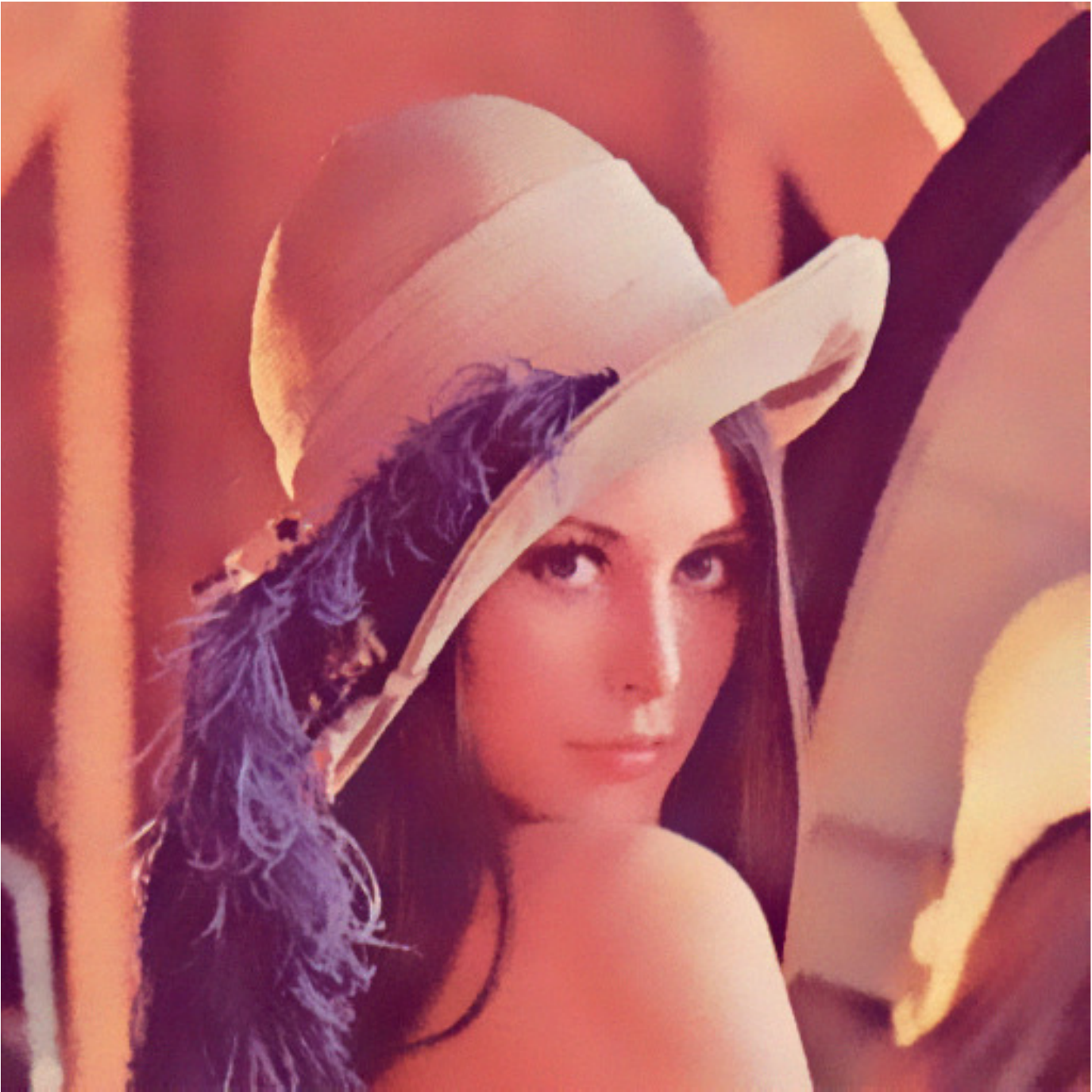}} \hspace{0.5mm}
\subfloat[AM ($5.8$, $29.4$, $0.87$).]{\includegraphics[width=0.19\linewidth,height=0.18\linewidth]{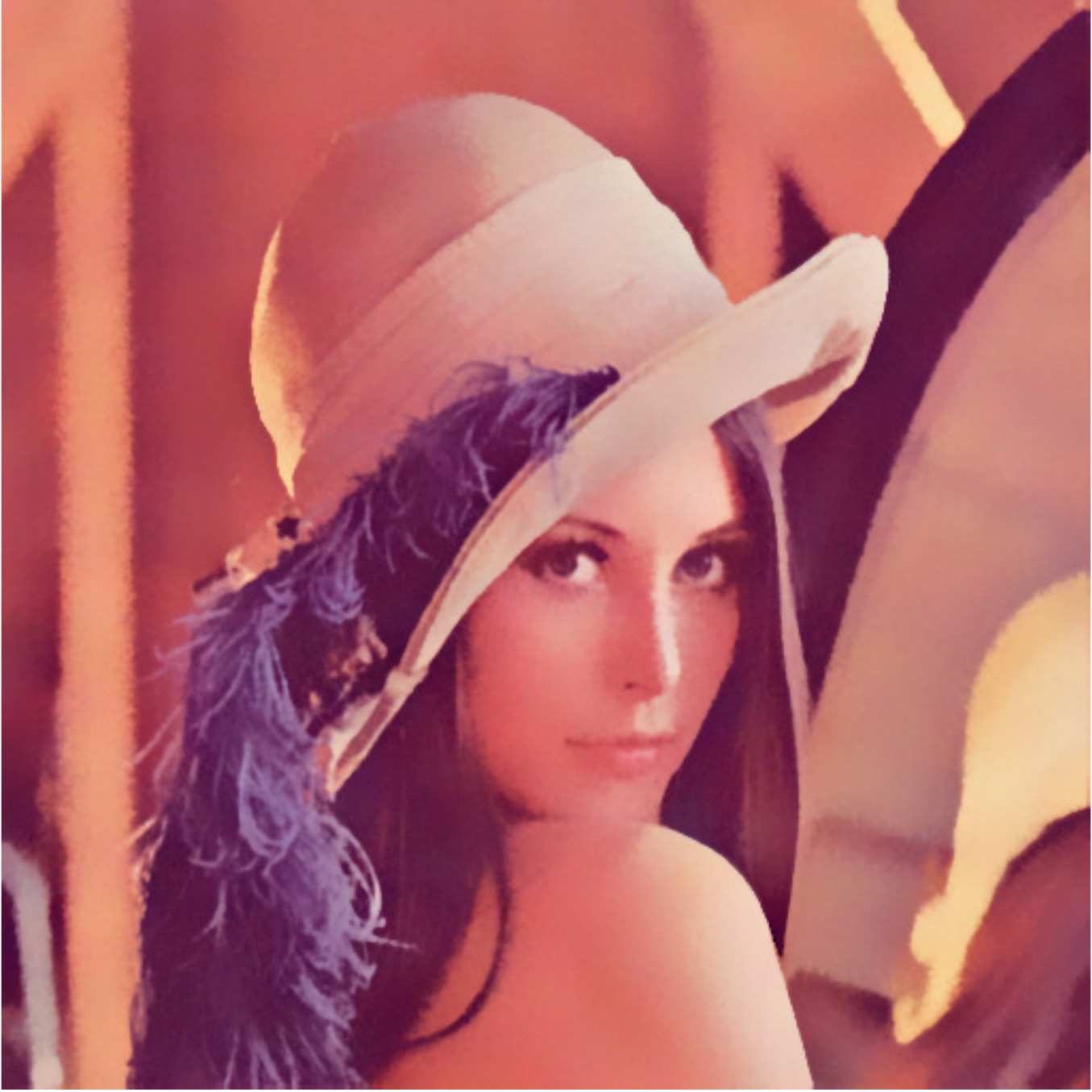}} \hspace{0.5mm}
\subfloat[BM3D ($5$, $33.01$, $0.93$).]{\includegraphics[width=0.19\linewidth,height=0.18\linewidth]{Lenadenoisedproposed-eps-converted-to.pdf}} 
\caption{Gaussian denoising (noise level $25/255$) of a color image using (b) PCA-NLM, its fast approximations (c) and (d), and (e) BM3D. The respective (Timing (sec), PSNR (dB), SSIM) is shown in the caption.}
\label{DenoiseNLM}
\end{figure*} 
\begin{figure}
\centering
\subfloat[Input ($256 \times 256$).]{\includegraphics[width=0.31\linewidth,height=0.28\linewidth]{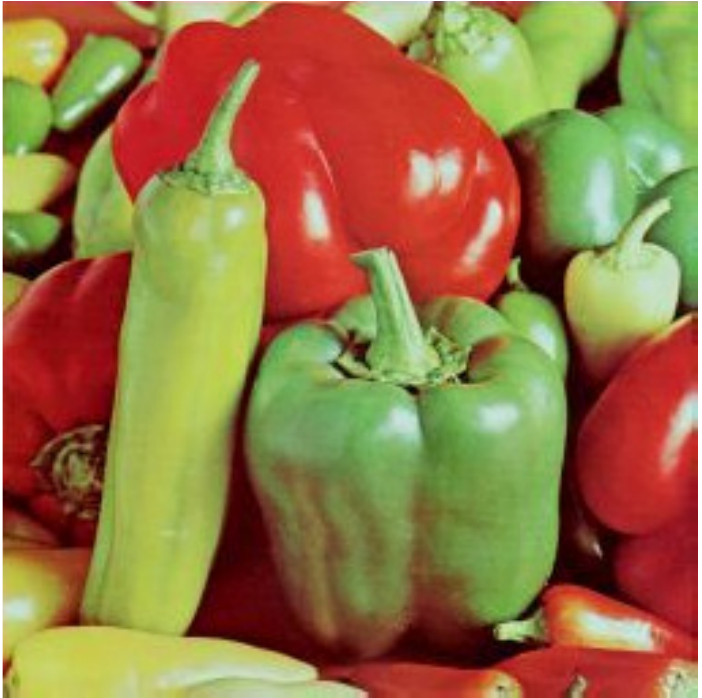}} \hspace{0.8mm}
\subfloat[\textbf{Ours ($108, 48.4$)}.]{\includegraphics[width=0.31\linewidth,height=0.28\linewidth]{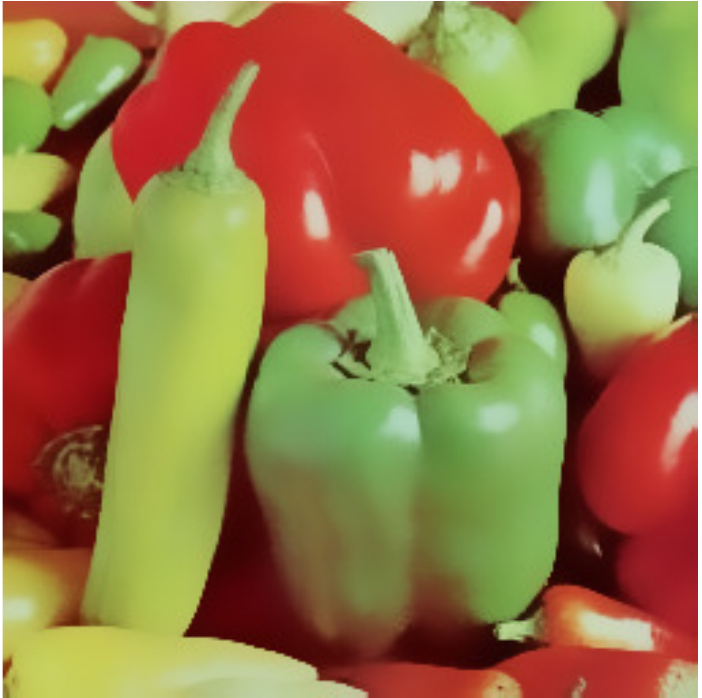}}\hspace{0.8mm}
\subfloat[\cite{mozerov2015global} ($107$, $46.5$).]{\includegraphics[width=0.31\linewidth,height=0.28\linewidth]{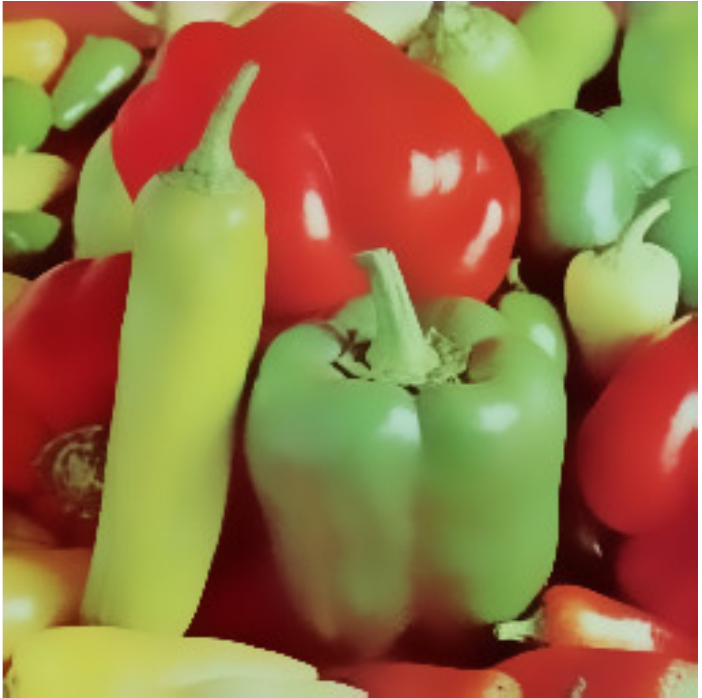}}

\subfloat[Brute-force, $4$ min.]{\includegraphics[width=0.31\linewidth,height=0.28\linewidth]{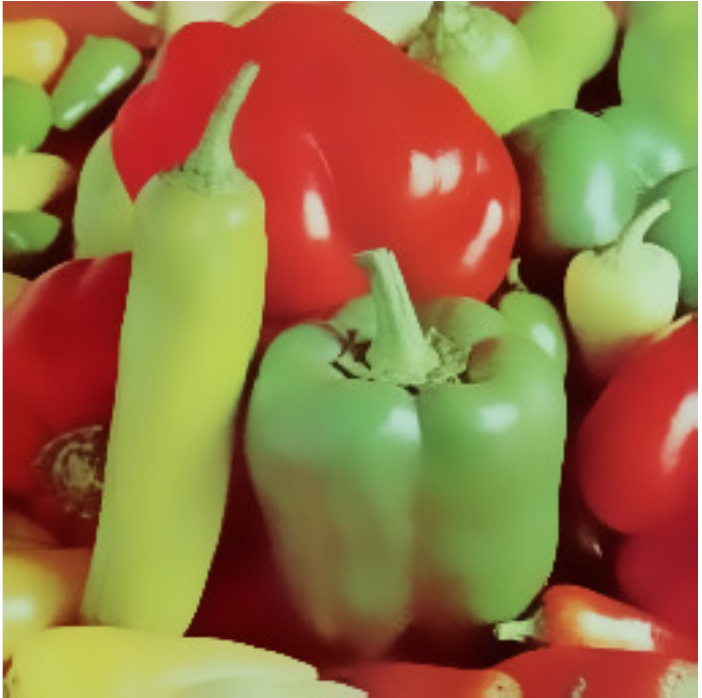}} \hspace{0.8mm}
\subfloat[\cite{gastal2012adaptive}, ($212$, $38.7$).]{\includegraphics[width=0.31\linewidth,height=0.28\linewidth]{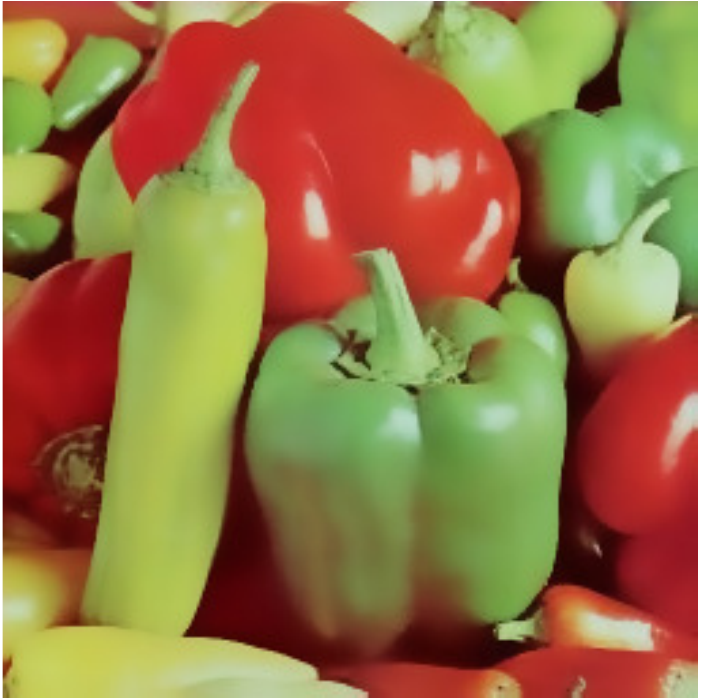}}\hspace{0.8mm}
\subfloat[\cite{adams2010fast}, ($44.5$).]{\includegraphics[width=0.30\linewidth,height=0.28\linewidth]{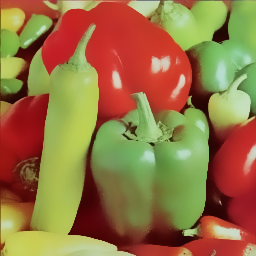}}
\caption{Visual comparison for fast BLF at  $\sigma=5$, $\theta=50$, and $|{\Omega}_0|=(6\sigma+1)^2$. The (timing, PSNR) are mentioned, where timing is in milliseconds and  PSNR is in dB. Timing is not mentioned for \cite{adams2010fast} which is implemented in C++. 
The breakup of timing for the proposed method is as follows: clustering ($11$ ms), eigendecomposition ($1$ ms), and convolutions ($96$ ms). Note that the overall timing is dominated by the convolutions. } 
\label{Colorbil}
\end{figure} 
\begin{figure}
\centering
\subfloat[Clean/Noisy.]{\includegraphics[width=0.45\linewidth,height=0.57\linewidth]{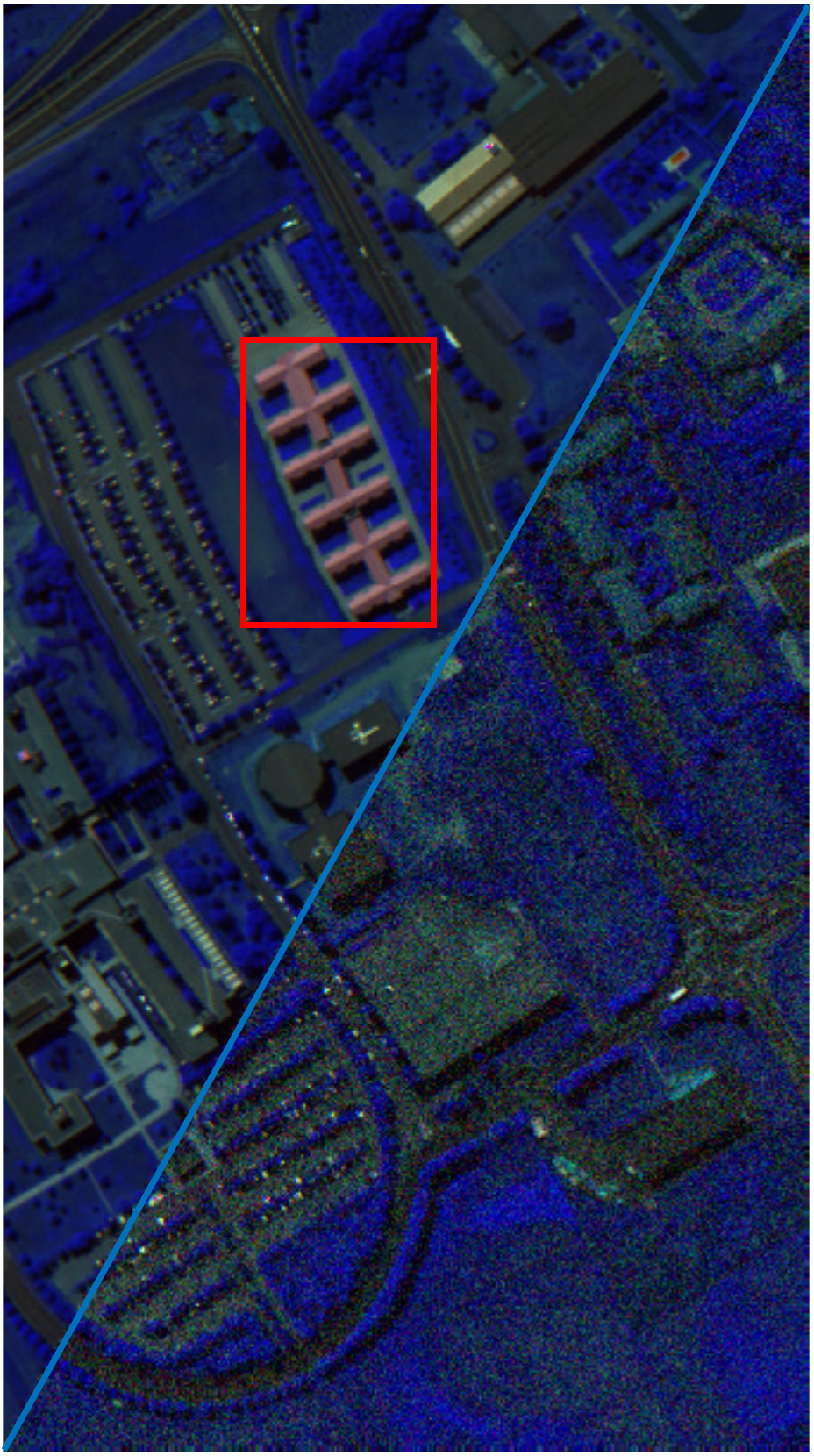}} \hspace{1mm}
\subfloat[\textbf{Ours} ($37$ sec, $31.2$ dB, $0.87$). ]{\includegraphics[width=0.45\linewidth,height=0.57\linewidth]{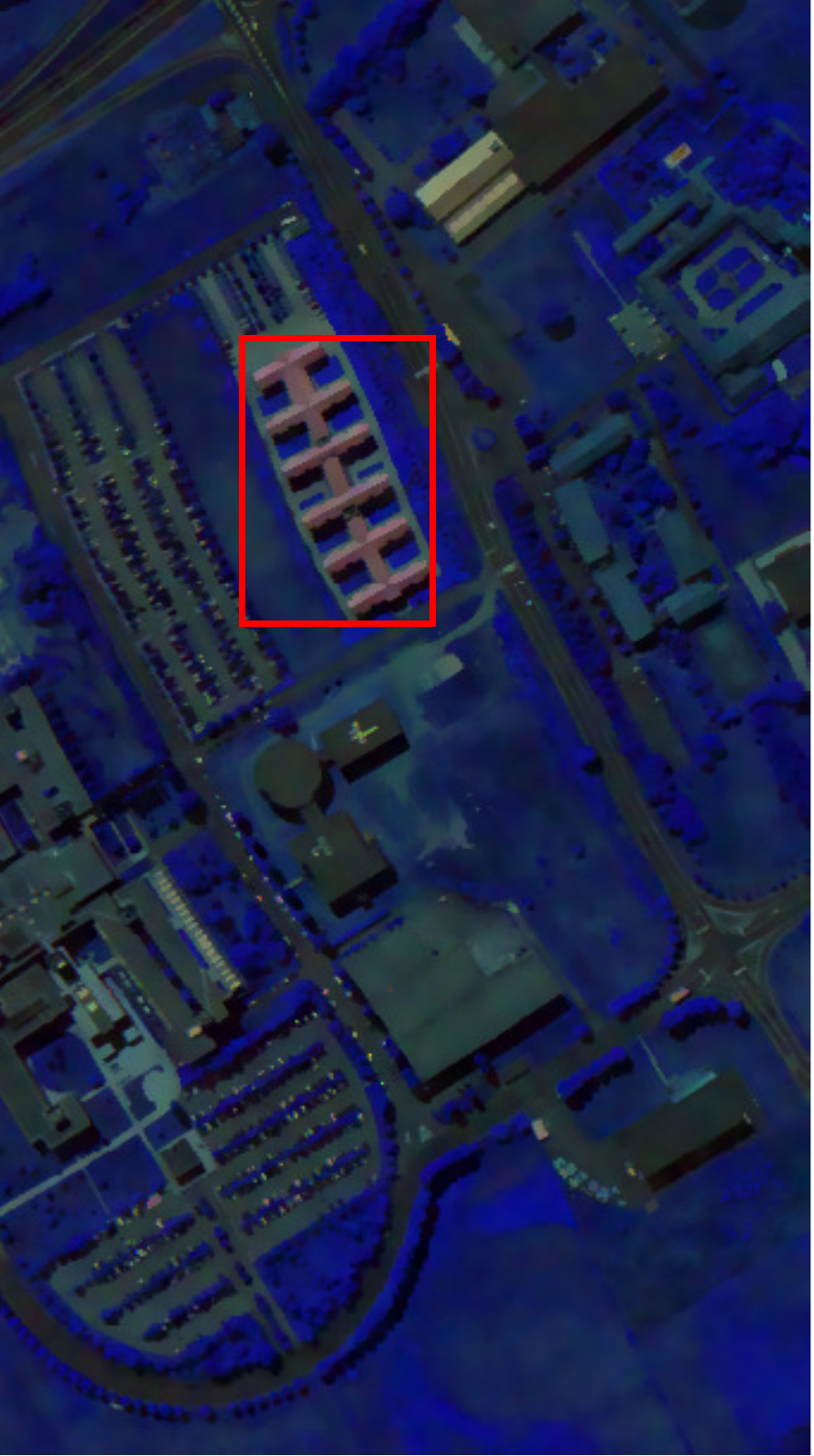}}

\subfloat[\cite{fan2017hyperspectral} ($3$ min, $30.54$ dB, $0.89$).]{\includegraphics[width=0.45\linewidth,height=0.57\linewidth]{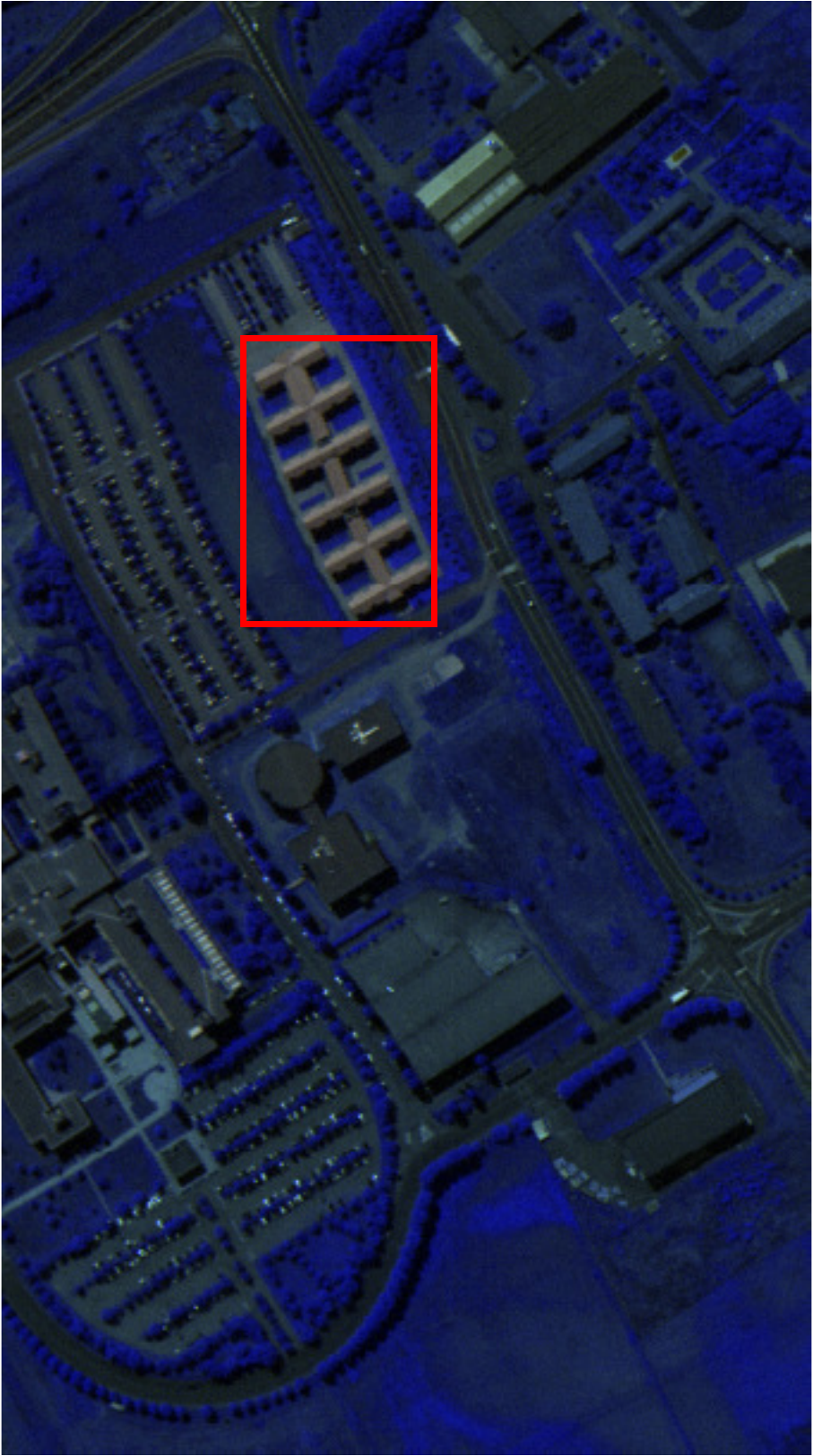}} \hspace{1mm}
\subfloat[\cite{zhao2015hyperspectral} ($20$ min, $30.09$ dB, $0.74$).]{\includegraphics[width=0.45\linewidth,height=0.57\linewidth]{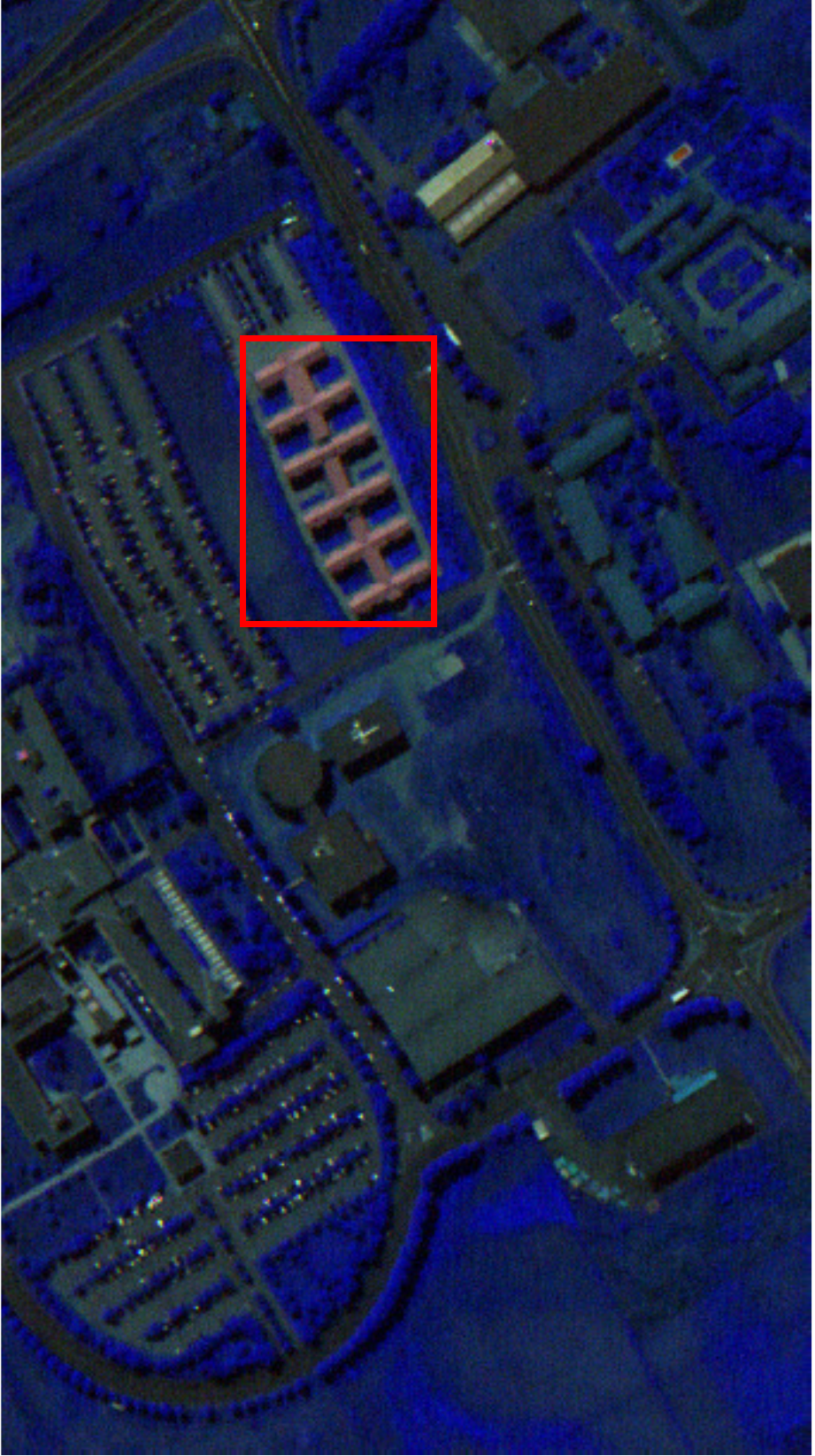}} 
\caption{Hyperspectral denoising of a natural image corrupted with Gaussian noise of level $25/255$. 
(Timing, PSNR, SSIM) are shown for all methods.}\label{Hyperspectral1}
\label{Hyperspectral2}
\end{figure} 

\textbf{Color BLF}. The state-of-the-art fast algorithms for color BLF are Adaptive Manifolds (AM) \cite{gastal2012adaptive}, Permutohedral Lattice (PL) \cite{adams2010fast}, and Global Color Sparseness (GCS) \cite{mozerov2015global}. 
We have compared with them in Figure \ref{Colorbil}. The number of manifolds is set automatically in AM, whereas we have used $15$ clusters in GCS and for the Nystr$\ddot{\text{o}}$m approximation. 
Following \cite{gastal2012adaptive,mozerov2015global}, we used PSNR to measure the error between the brute-force and fast implementations. In Figure \ref{Colorbil}, notice that while our PSNR marginally exceeds that of GCS, it is however much better than PL and AM.  Also notice the significant acceleration over the brute-force implementation obtained using our algorithm. 
We have also provided  a table comparing the different methods on the Kodak dataset \cite{ImageSource2} in the supplement.
The table shows that our method is better than GCS and PL when $\theta > 40$. 
As claimed in the introduction, we can see from the table that clustering provides a significant boost in filtering accuracy ($10\mbox{-}20$ dB) over uniform sampling. 

\textbf{Color NLM}. AM is the state-of-the-art fast algorithm for color NLM (and PCA-NLM). In NLM, $\rho=3(2r+1)^2$, where $r$ is the patch radius \cite{buades2005non}. On the other hand,  $\rho$ is reduced to a smaller value in PCA-NLM using PCA. 
Following \cite{tasdizen2009principal}, we set $\theta$ to be three times the noise level for all experiments. 
Denoising results are shown in Figure \ref{DenoiseNLM}, where $S=10$ and $r=3$. 
For (b), (c), and (d), PCA was used to reduce the range dimension from $3 \times 7^2$ to $25$. 
We used $31$ clusters (resp. manifolds) for the Nystr$\ddot{\text{o}}$m approximation (resp. AM). 
Following \cite{wang2004}, we measured the denoising performance using PSNR and SSIM (between the clean and denoised images). 
Note that we are superior to AM both in terms of accuracy and timing. 
Importantly, our PSNR is close to PCA-NLM (the method being approximated), but we are about $160\times$ faster. 
In comparison with BM3D \cite{dabov2006image}, our PSNR is $3$ dB less. However, our timing is about half that of BM3D, since our complexity is much less than that of BM3D. 
Additional visual comparisons and accuracy analysis is provided in the supplement. 

\textbf{Hyperspectral BLF}. 
Finally, we present a denoising result for a hyperspectral image of size $(610 \times 340) \times 103$ bands using BLF ($\sigma=3, \theta=100$).
We have also compared with state-of-the-art methods for hyperspectral denoising \cite{fan2017hyperspectral,zhao2015hyperspectral}, 
whose parameters have been tuned accordingly. The results are shown in Figure \ref{Hyperspectral2}. 
We have used $m_0=32$ landmarks for the Nystr$\ddot{\text{o}}$m approximation. 
As a standard practice, the $\mathrm{PSNR}$ and SSIM values are averaged over the spectral bands. 
Notice that our method can restore details better, which results in higher PSNR/SSIM values. In particular, the color is not satisfactorily restored in \cite{fan2017hyperspectral} and grains can be seen in \cite{zhao2015hyperspectral}. 
Being a one-shot method, we are much faster than \cite{fan2017hyperspectral,zhao2015hyperspectral}.

\section{Conclusion}
\label{Conc}

We showed that fast bilateral and nonlocal means filtering of high-dimensional images can be performed using the Nystr$\ddot{\text{o}}$m approximation. The proposed algorithm can significantly accelerate the brute-force implementation of these filters, without compromising the visual quality. In particular, our algorithm is often competitive with state-of-the-art fast algorithms, and comes with provable guarantee on the filtering accuracy.

\bibliographystyle{IEEEtran}
\bibliography{citations}

\end{document}


\title{\qquad \large{Supplementary for ``Fast High-Dimensional Kernel Filtering''}}
\author{\large{P. Nair and K. N. Chaudhury}}
\date{}

\maketitle

\begin{figure}
\centering
\subfloat[\tiny{Clean/Noisy ($12$ dB)}.]{\includegraphics[width=0.3\linewidth]{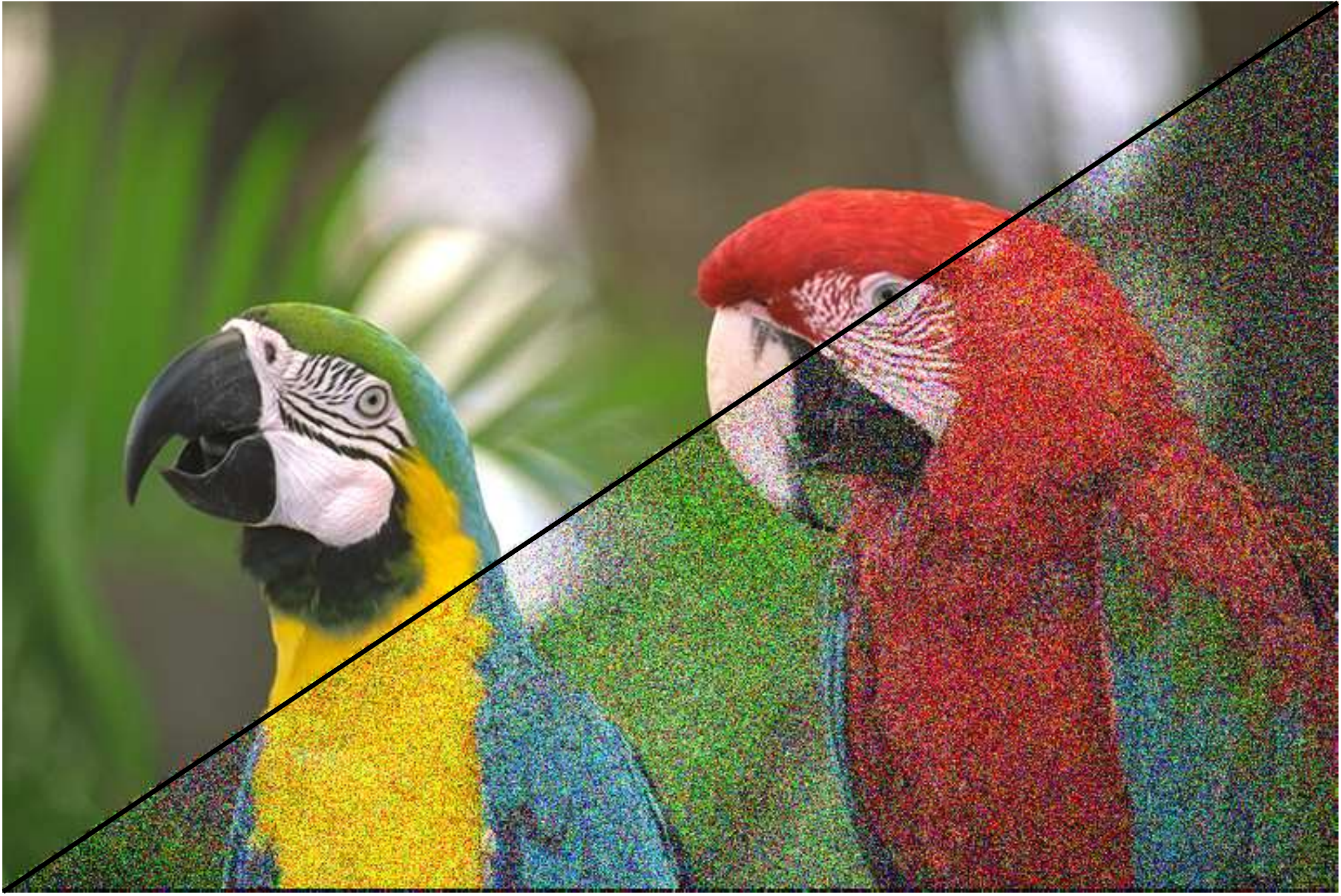}} \hspace{0.1mm}
\subfloat[\tiny{NLM ($26.7$ dB, $10$ min)}.]{\includegraphics[width=0.3\linewidth]{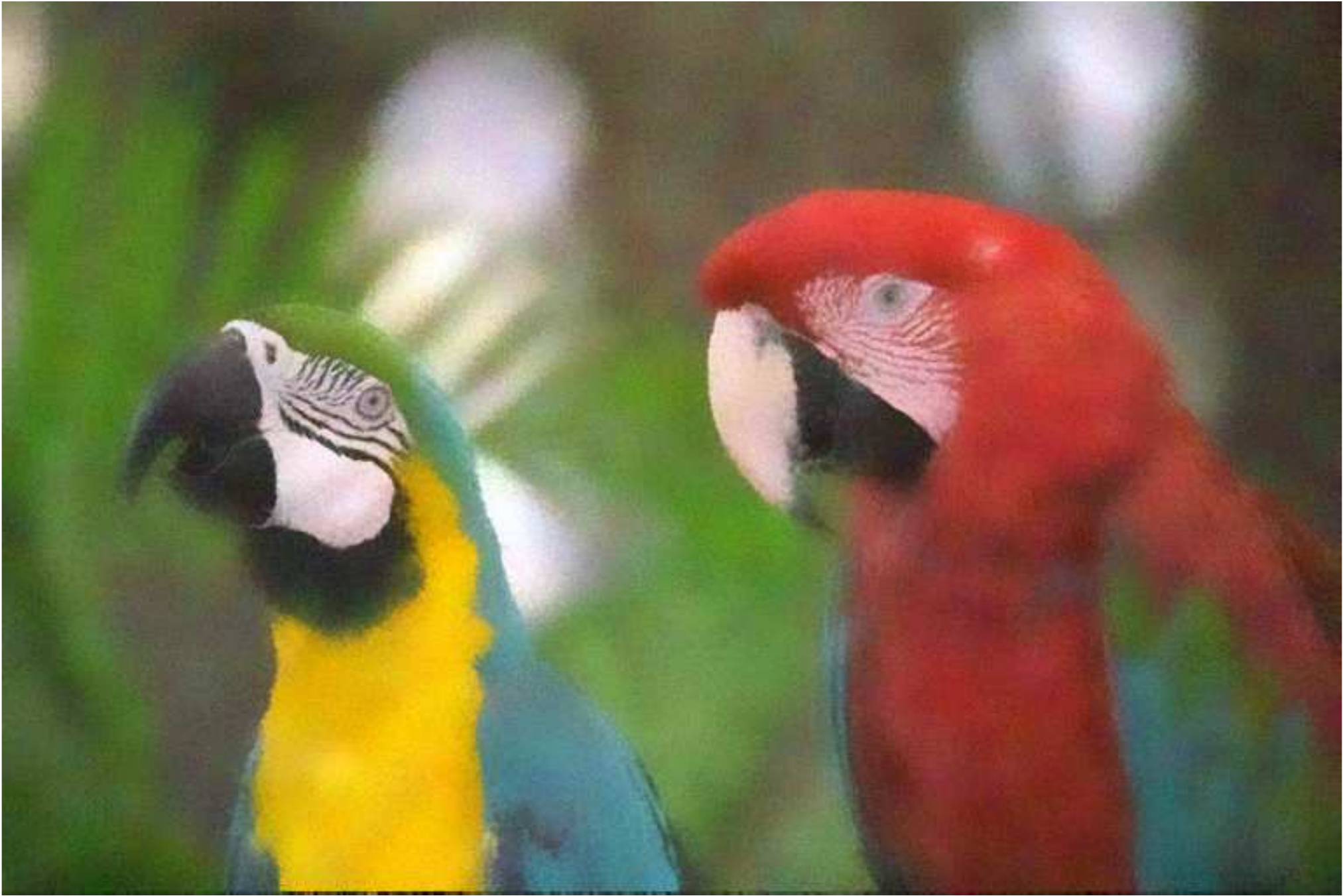}} 

\subfloat[\textbf{\tiny{Ours ($\textbf{26.9}$ dB, $\textbf{2.7}$ sec)}}.]{\includegraphics[width=0.3\linewidth]{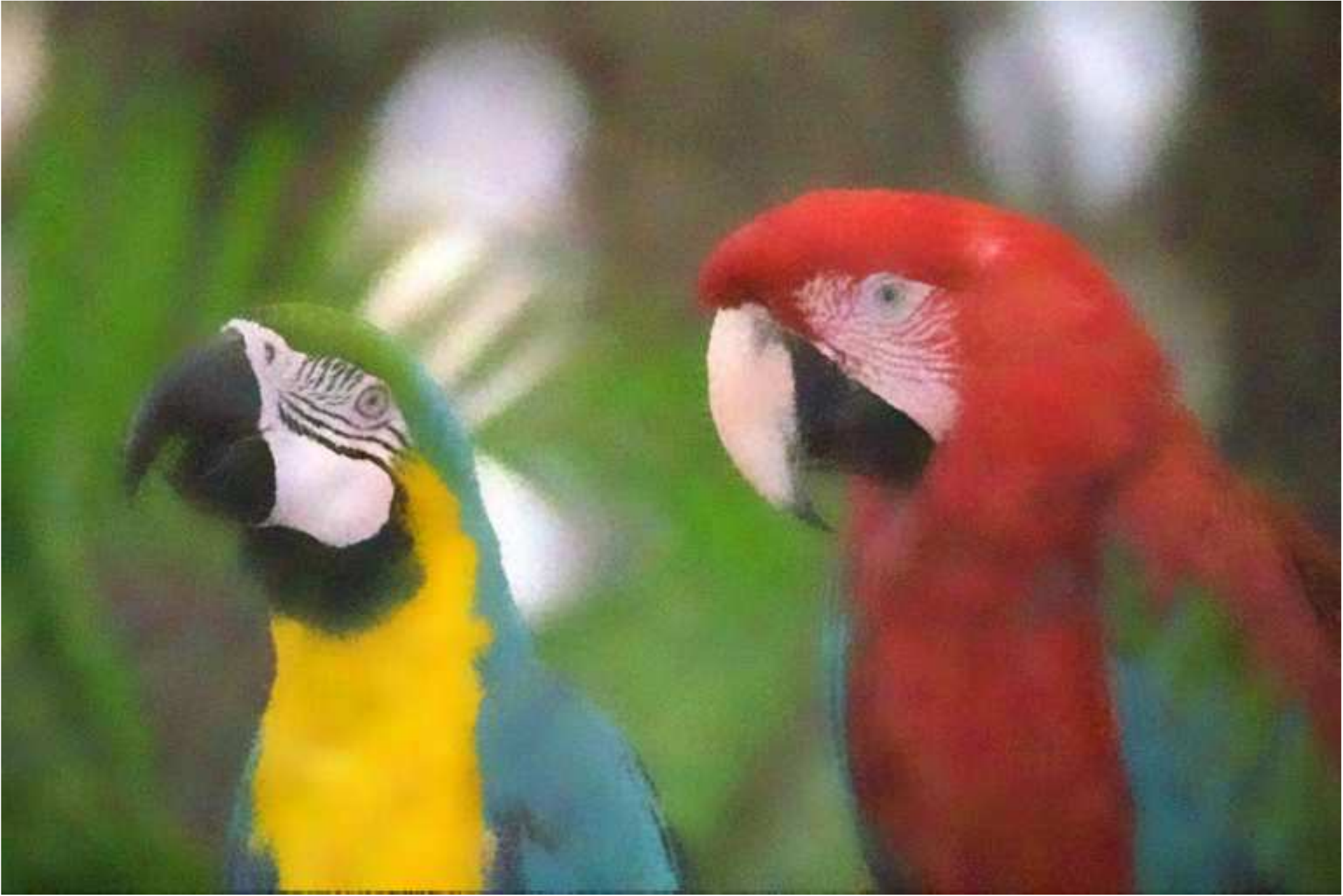}} \hspace{0.1mm}
\subfloat[\tiny{AM ($25$ dB, $4.2$ sec)}.]{\includegraphics[width=0.3\linewidth]{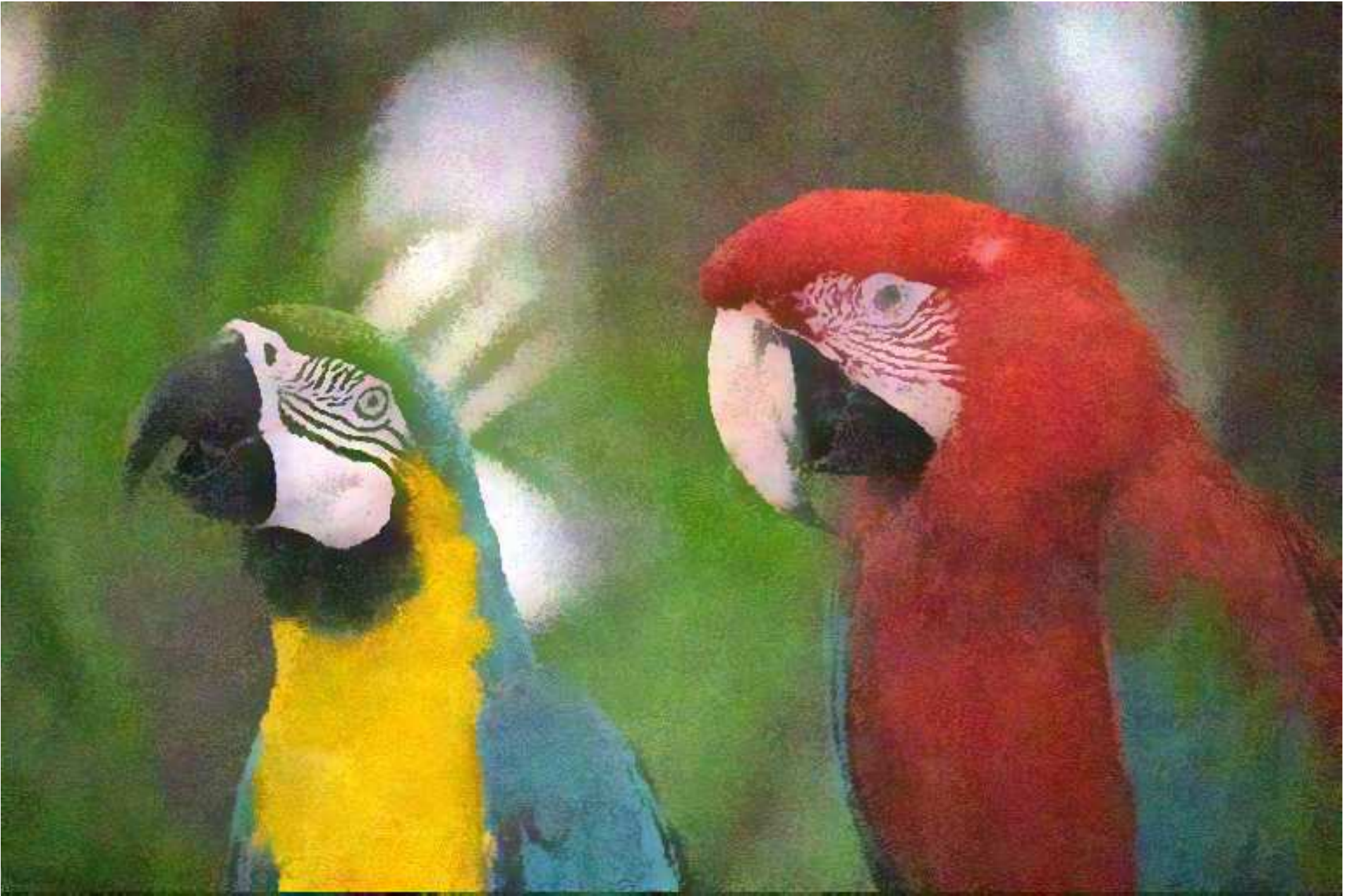}} \hspace{0.1mm}
\subfloat[\tiny{BM3D ($30.7$ dB, $15.4$ sec)}.]{\includegraphics[width=0.3\linewidth]{kodim23BM3D.eps}} 
\caption{Gaussian denoising of a color image (noise level $63/255$). Our PSNR is almost identical to that of PCA-NLM,  and we outperform AM by a considerable margin.
Parameters: $m_0 = 31, S = 10, r = 3$, and the PCA dimension is $25$.}
\end{figure}

\begin{figure}
\centering
\includegraphics[width=0.7\linewidth]{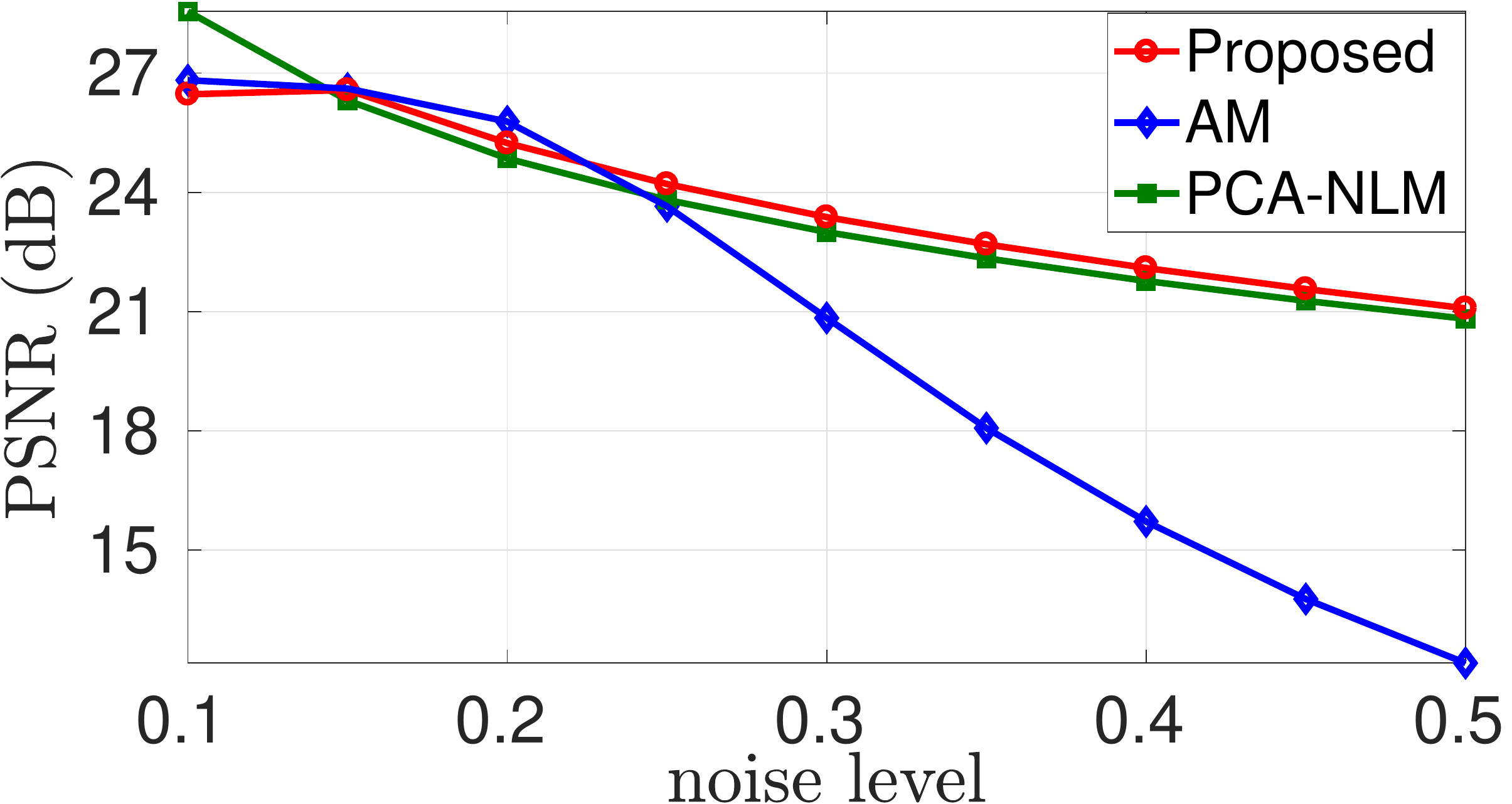}
\caption{Denoising performance of PCA-NLM, AM and our method at various noise levels ($\times 255$). 
Parameters: $m_0 = 31, S = 10, r = 3$, and the PCA dimension is $6$.
The $\mathrm{PSNR}$ is averaged over the color images in the Kodak dataset. 
Notice that we outperform AM by a large margin at large noise levels.
}
\end{figure}

\begin{table}
\centering
\setlength{\tabcolsep}{0.5pt}
\centering
\label{table2}
\begin{tabular}{| c | c  c c c c c c|}
 \hline
Method \textbackslash{} $(\sigma,\theta)$   & $(5, 30)$  & $(5, 50)$  & $(5, 60)$  &  $(10, 30)$  &  $(10, 50)$ &  $(10, 60)$ &\\ \hline
AM [$20$]    &44.5  &43.6  &43.4  &42.5 &41.4 &41.1& \\ \hline
[$19$]  &50.2  &47.7  &46.8  &\textbf{49.6}  &46.5  &45.6& \\ \hline
GCS [$21$]  &\textbf{50.3}  &50.2  &49.6  &48.5  &47.9  &47.3& \\ \hline
\textbf{Ours  (uniform)}   	 &20.3  &36.1  &41.2  &21.1  &30.4  &37.2& \\ \hline
\textbf{Ours   ($k$-means)} &40.7  &\textbf{50.5}  &\textbf{53.9}  &38.2  &\textbf{49.1}  &\textbf{53.2}& \\ \hline
\end{tabular}
\caption{Average $\mathrm{PSNR}$ for bilateral filtering of color images from the Kodak dataset. 
We used $m_0=15$ clusters for our method and GCS, and the default settings for AM.
Notice that our method performs better than GCS and AM for large $\theta$. 
The important observation is that there is a massive boost in accuracy ($10$ to $20$ dB) using $k$-means over uniform sampling. }
\end{table} 

\noindent \textbf{Proof of Theorem $2$} (sketch): We write equations (1) and (13) as $\g(\x) = \boldsymbol{\zeta}(\x)/\eta(\x)$ and  $\hat\g(\x) = \hat{\boldsymbol{\zeta}}(\x)/\hat\eta(\x)$,
where $ \boldsymbol{\zeta}(\x)$ and $\hat{\boldsymbol{\zeta}}(\x)$ are the respective numerators, and $\eta(\x)$ and $\hat\eta(\x)$ are the  respective denominators. Note that 
\begin{align*}
\hat\g(\x) - \g(\x) = \eta(\x)^{-1}\Big(\hat{\g}(\x) \big( \eta(\x)-\hat\eta(\x) \big) + \big(\hat{\boldsymbol{\zeta}}(\x) - \boldsymbol{\zeta}(\x)\big)\Big).
\end{align*}
Using the triangle inequality, we can bound ${\lVert{\hat\g}(\x) - \g(\x)\rVert}$ by
\begin{equation}
\label{main}
\frac{1}{\lvert\eta(\x)\rvert} \Big(\sqrt{nR}{\lvert \eta(\x)-\hat\eta(\x) \rvert}  + {\lVert \hat{\boldsymbol{\zeta}}(\x) - \boldsymbol{\zeta}(\x) \rVert} \Big),
\end{equation}
\begin{equation*}
\text{Note that} \ \ \hat{\boldsymbol{\zeta}}(\x)-\boldsymbol{\zeta}(\x)=\sum_{\y \in W_{\x}} \omega(\x - \y)  \Delta(\x,\y)  \f(\y),
\end{equation*}
\begin{equation*}
\text{where} \ \ \Delta(\x,\y) = \widehat\K(\iota(\x),\iota(\y)) - \K(\iota(\x),\iota(\y)).
\end{equation*}
In particular, $\lvert \Delta(\x,\y) \rvert \leq \lVert \widehat\K  - \K \rVert_{\text{F}}$ for any $\x,\y \in \Omega$. Therefore, 
\begin{equation}
\label{numbound}
{\lVert \hat{\boldsymbol{\zeta}}(\x)-\boldsymbol{\zeta}(\x) \rVert} \leq \omega(\boldsymbol{0}) \lvert W_{\x} \rvert \sqrt{nR} \lVert \widehat\K  - \K \rVert_{\text{F}}, 
\end{equation}
since ${\lvert \omega(\x - \y) \rvert} \leq \omega(\boldsymbol{0})$ and  ${\lVert \f(\y) \rVert} \leq \sqrt{n}$. Similarly,
\begin{equation}
\label{denbound}
{\lvert \eta(\x)-\hat\eta(\x) \rvert} \leq \omega(\boldsymbol{0}) (2S+1)^d \lVert\widehat\K  - \K \rVert_{\text{F}},
\end{equation}
since $\lvert W_{\x} \rvert=(2S+1)^d$. Now, as $\omega$ and $\kappa$ are positive, $ \eta(\x)  \geq \omega(\boldsymbol{0}) \kappa(\boldsymbol{0},\boldsymbol{0})$. Combining the above bounds, we obtain
\begin{equation*}
\lVert \hat\g(\x) - \g(\x)  \rVert \leq \frac{2}{\kappa(\boldsymbol{0},\boldsymbol{0})} (2S+1)^d \sqrt{nR} \lVert \widehat\K  - \K \rVert_{\text{F}}.
\end{equation*}
Along with Proposition $1$ (in the manuscript), we conclude that
\begin{equation*}
\max_{\x \in \Omega} \  \lVert \hat\g(\x) - \g(\x) \rVert  \leq C_1 \sqrt{e} + C_2 e,
\end{equation*}
where $C_i= (2/\kappa(\boldsymbol{0},\boldsymbol{0}))  (2S+1)^d \sqrt{nR} c_i$ for $i=1,2$.

\begin{figure}
\centering
\subfloat[Ours (107,48.4).]{\includegraphics[width=0.23\linewidth]{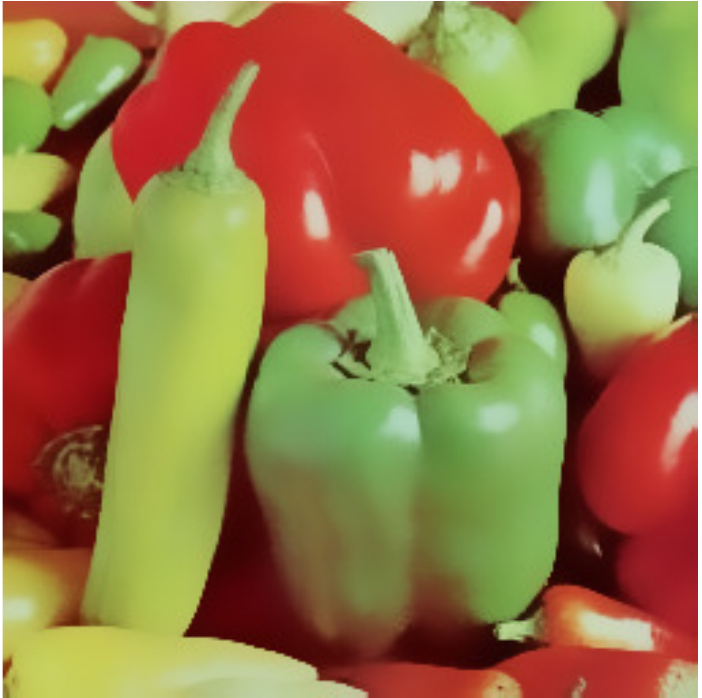}} \hspace{0.1mm}
\subfloat[{[$12$] ($1050$, $40.9$)}]{\includegraphics[width=0.23\linewidth]{peppersIJCV.eps}} \hspace{0.1mm}
\subfloat[{[$27$] ($960$, $30.2$)}]{\includegraphics[width=0.23\linewidth]{peppersVV.eps}} \hspace{0.1mm}
\subfloat[{[$28$] ($970$, $35.8$)}]{\includegraphics[width=0.23\linewidth]{peppersSBF.eps}} 
\caption{Comparison with [$12,27,28$] for color bilateral filtering. The timings are in milliseconds. 
The number of convolutions used for our method, [$12$], [$27$] and [$28$] are $30, 128, 200$, and $200$, respectively.
Even with fewer  convolutions, our PSNR is better by $8$ to $18$ dB.}
\end{figure}
\begin{figure}
\centering
\includegraphics[width=0.75\linewidth]{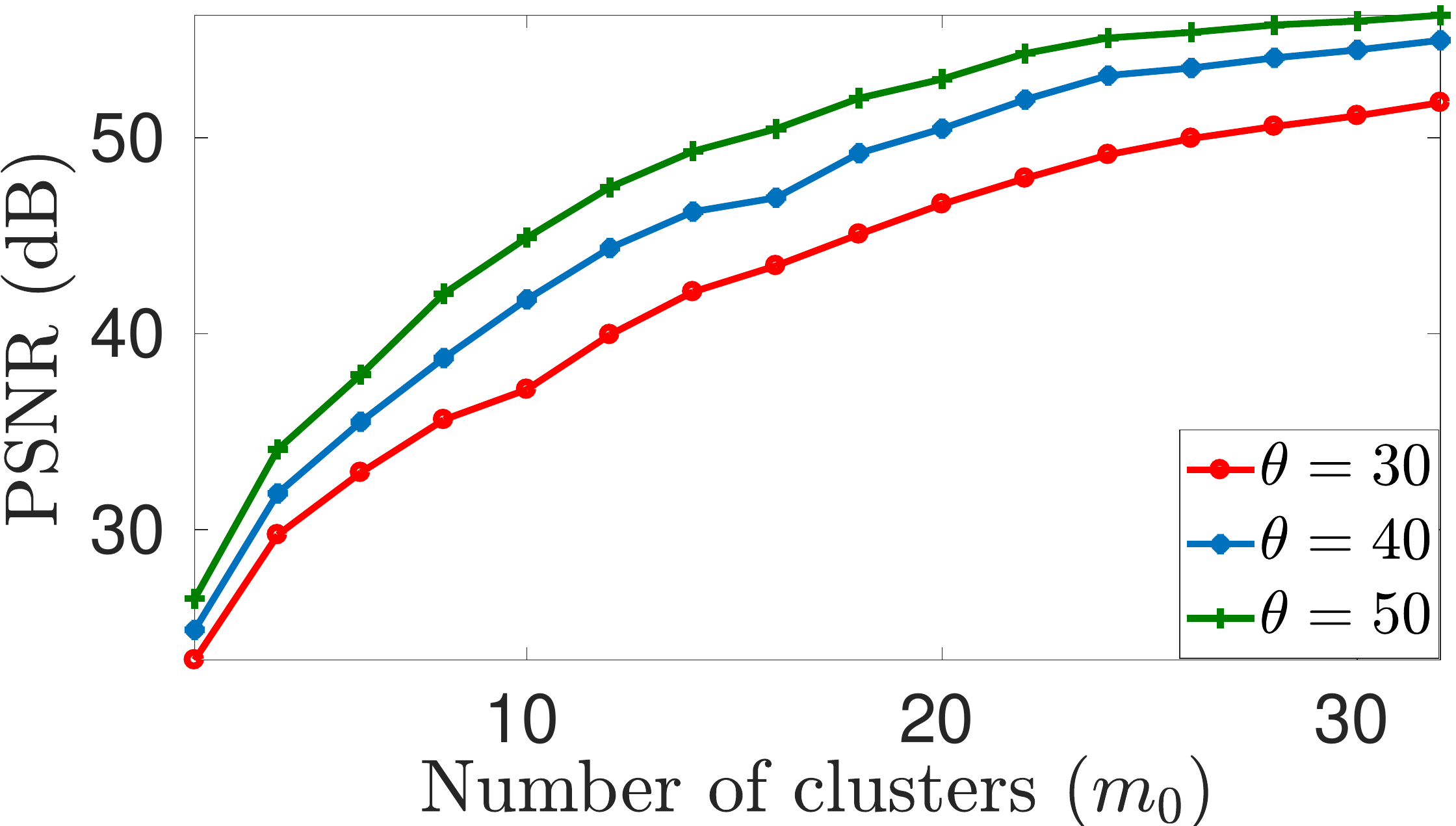}
\caption{$\mathrm{PSNR}$ vs $m_0$ for color bilateral filtering for varying $\theta$ and fixed $\sigma=10$. The $\mathrm{PSNR}$ is averaged over the color images in the Kodak dataset. 
As supported by Theorem $2$, the accuracy increases with the number of clusters.}
\end{figure}